\documentclass[11pt]{article}
\usepackage{colacl06}
\usepackage{times}
\usepackage{latexsym}
\usepackage{xspace}
\usepackage{graphicx}
\usepackage{amssymb}
\usepackage{amsmath}
\usepackage{soul}
\usepackage{cancel}
\usepackage[usenames,dvipsnames]{color} 

\newcommand{\omt}[1]{}

\newcommand{\debate}{debate\xspace}
\newcommand{\debates}{{\debate}s\xspace}

\newcommand{\spunit}{speech segment\xspace}
\newcommand{\spunith}{speech-segment\xspace}

\newcommand{\spunits}{speech segments\xspace}

\newcommand{\y}{``yea''\xspace}
\newcommand{\n}{``nay''\xspace}

\newcommand{\refthres}{\theta_{\mbox{agr}}\xspace} 
\newcommand{\refscore}{\mathit{agr}\xspace}
\newcommand{\refcoef}{\alpha\xspace}
\newcommand{\refstd}{\sigma_{\spvar}}
\newcommand{\refmean}{\mu\xspace}
\newcommand{\yclass}{{\cal Y}\xspace}
\newcommand{\nclass}{{\cal N}\xspace}
\newcommand{\class}{C}
\newcommand{\classfn}{c}
\newcommand{\indscore}{\mbox{{\it ind}}\xspace}
\newcommand{\spvar}{s}
\newcommand{\refconfvar}{d}
\newcommand{\refconf}{\refconfvar(\spvar)\xspace} 
\newcommand{\refvar}{r}
\newcommand{\agrconfvar}{d}
\newcommand{\agrconf}{\agrconfvar(\refvar)}
\newcommand{\agrstd}{\sigma_\refvar}
\newcommand{\numspunits}{n}
\newcommand{\link}{\ell}
\newcommand{\strength}{{\mathit str}}

\newcommand{\resmajoritybl}{majority baseline\xspace}
\newcommand{\reswordbl}{$\#(\mbox{``support''}) - \#(\mbox{``oppos''})$\xspace}
\newcommand{\ressvmbl}{SVM\xspace}
\newcommand{\spaceplus}{+~}
\newcommand{\resspeakerlink}{SVM + same-speaker links\xspace}

\newcommand{\refs}{agreement links\xspace}
\newcommand{\useposref}{$\refthres = 0$}
\newcommand{\useavgref}{$\refthres = \refmean$}
\newcommand{\resfullref}{\spaceplus \refs,  \useposref \xspace}
\newcommand{\reslimitedref}{\spaceplus \refs, \useavgref \xspace}

\newcommand{\amendabbrev}{amdmts\xspace}
\newcommand{\agrSVMtrainnoamend}{\mbox{Train: no \amendabbrev}; \useposref}
\newcommand{\agrSVMtrainwamend}{\mbox{Train: with \amendabbrev}; \useposref}

\newcommand{\Localclass}{Segment-based\xspace}

\newcommand{\Localsvmsetting}{\Localclass \spunith
classification\xspace}
\newcommand{\globalclass}{speaker-based\xspace}
\newcommand{\Globalclass}{Speaker-based\xspace}

\newcommand{\Globalsvmsetting}{\Globalclass \spunith classification\xspace}

\newcommand{\bestdev}[1]{{\bf #1}}
\newcommand{\testbestdev}[1]{{\bf #1}}

\newcommand{\threshstrike}[1]{\st{#1}}
\newcommand{\fix}[2]{{\color{Red}{\threshstrike{#1}~#2}}}

\author{Matt Thomas, Bo Pang, and Lillian Lee \\
Department of Computer Science, Cornell University \\
Ithaca, NY 14853-7501 \\
{ {mattthomas84@gmail.com}, {pabo@cs.cornell.edu},
  {llee@cs.cornell.edu}}\\ \\ 
\fbox{\parbox{6in}{Original version:
    Proceedings of EMNLP
  2006, pp. 327--335. This revision of Dec 2006 \space includes minor updates to
  four data points that were originally slightly off due to a single
  incorrect threshold value. All changes
  are indicated by both the color {\color{Red}{red}} and by
  \threshstrike{strikethroughs} or [bracketed comments], except for
  the reformatting of the bibliography.}}}

\title{Get out the vote: Determining  support or opposition from 
  Congressional floor-debate transcripts}

\begin{document}
\maketitle

\begin{abstract}

We investigate whether one can determine from the transcripts of
U.S. Congressional floor debates whether the 
speeches 
represent support of or opposition to proposed legislation.  To address this
problem, we exploit 
the fact that these speeches occur as part of a discussion; this allows us to use
sources of information regarding relationships
between 
discourse segments,
such as whether a given utterance indicates agreement with the
opinion expressed by another.  We find that the incorporation of such information
yields
substantial
improvements over 
classifying speeches in isolation.

\end{abstract}


\section{Introduction}
\label{sec:intro}

\begin{quote}
{\em One ought to recognize that the present political chaos is connected
with the decay of language, and that one can probably bring about some
improvement by starting at the verbal end.}  --- Orwell,
``Politics and the English language''
\end{quote}

\medskip

We have entered an era where very large amounts of politically
oriented text are now available online.
This
includes both official documents, such as the full text of 
laws and the proceedings of legislative bodies, and unofficial
documents, such as postings on weblogs (blogs) devoted to politics.
In some sense, 
the availability of such data is simply a manifestation of 
a general trend of ``everybody putting their records on the
Internet''.\footnote{It is worth pointing out that the United States'
Library of Congress was an extremely early adopter of Web technology:
the THOMAS database (http://thomas.loc.gov) of congressional bills and related data was
launched in January 1995, when Mosaic was not quite two years old and
Altavista did not yet exist.}
The online accessibility of 
politically oriented texts in particular, however,
is a phenomenon that some have gone so far as to say will have a potentially
society-changing effect.

In the 
United States, 
for example,
governmental bodies are providing and soliciting political
documents via the Internet,
with lofty goals in mind: 
{\em electronic rulemaking} (eRulemaking)
initiatives
involving the
``electronic collection,
distribution, synthesis, and analysis of public commentary in the
regulatory rulemaking process'', may ``[alter] the citizen-government
relationship'' \cite{Shulman+Schlosberg:02a}.  Additionally, much media attention
has been focused  recently on the potential impact that
Internet sites 
may have on 
politics\footnote{E.g., ``Internet injects sweeping change
into U.S. politics'', Adam Nagourney, {\em The New York Times}, April
2, 2006.}, or at least on political journalism\footnote{E.g., ``The
End of News?'', Michael Massing, {\em The New York Review of Books}, 
December 1, 2005.}. 
Regardless of whether one views such claims as clear-sighted prophecy
or mere hype,
it is obviously important to help people understand and analyze
politically oriented text,
given the importance of enabling
informed participation in the political process.

Evaluative and persuasive documents,
such as a politician's speech regarding a bill or 
a blogger's commentary on a
legislative proposal, form a particularly interesting type of
politically oriented text.  
People are much more likely to consult such 
evaluative statements 
than the actual text of a bill or law
under discussion, 
given the dense nature of legislative language and the fact that (U.S.) bills
often reach several hundred pages in length
\cite{Smith+Robert+VanderWielen:05a}. Moreover, 
political opinions are explicitly solicited in the eRulemaking scenario.

In the analysis of 
evaluative language, it is fundamentally
necessary to determine whether the author/speaker supports or
disapproves of the topic of discussion.  
In this paper, we investigate
the following specific instantiation of this problem: 
we seek to determine
from the transcripts of U.S. Congressional floor debates whether each
``speech'' (continuous single-speaker segment of text) represents
support for or opposition to a proposed piece of legislation.  Note
that from an experimental point of view, this is a very convenient
problem to work with because we can automatically determine ground
truth (and thus avoid the need for manual annotation) simply by
consulting publicly available 
voting records.

\paragraph{
\mbox{Task properties}}

Determining whether or not a speaker 
supports
a 
proposal falls within the realm of {\em
sentiment analysis}, an extremely active research area devoted to the
computational treatment of subjective or opinion-oriented language
(early work includes Wiebe and Rapaport \shortcite{Wiebe+Rapaport:88a},
Hearst \shortcite{Hearst:92a}, Sack \shortcite{Sack:94a}, and \newcite{Wiebe:94a};
see Esuli \shortcite{Esuli:sentbib} for an active bibliography).  In
particular, since we treat 
each individual speech
within a 
debate as a single ``document'', 
we are considering 
a version of
{\em document-level sentiment-polarity classification}, 
namely,  automatically distinguishing between 
positive and negative
documents
\cite{Das+Chen:01a,Pang+Lee+Vaithyanathan:02a,Turney:02a,Dave+Lawrence+Pennock:03a}.

Most sentiment-polarity classifiers proposed in the recent literature  categorize
each document independently.  A few others  incorporate
various measures of inter-document 
similarity between the texts to be labeled
\cite{Agarwal+Bhattacharyya:05a,Pang+Lee:05a,Goldberg+Zhu:06a}.  
Many interesting 
opinion-oriented documents, however, can be linked through certain relationships 
that occur in the context of evaluative {\em discussions}.
For example, we may find textual\footnote{
Because we are most interested in techniques applicable across
domains, we
restrict consideration 
to 
NLP aspects of the problem, 
ignoring external problem-specific information.
For
example, although most votes 
in our corpus were almost completely along party lines (and despite
the fact that same-party information is easily incorporated via
the methods we propose), 
we did not use party-affiliation data.
Indeed, 
in
other settings (e.g., a movie-discussion listserv) one 
may not be able to determine the participants' political
leanings,  
and such information may not lead to significantly improved results
even if it were available.
}
evidence of a high likelihood of {\em agreement} between two speakers, 
 such as explicit assertions (``I second
that!'') or quotation of messages in emails or postings (see
\newcite{Mullen+Malouf:06a} but cf. \newcite{Agrawal+al:03a}).
Agreement evidence can be a powerful aid in our classification
task: for example,  we can
easily categorize a complicated (or overly terse) document if we find
within it indications of agreement with a clearly positive text.

Obviously, incorporating agreement information provides additional
benefit only when the 
input
documents 
are relatively difficult to classify
individually.  Intuition suggests that this is true of the
data with which we experiment, for several reasons.  
First, U.S. congressional debates contain
very rich language and cover an extremely wide variety of topics, 
ranging from flag burning to international policy to the federal budget. 
Debates are also subject to digressions, some fairly natural 
and others less so (e.g., ``Why
are we discussing this bill when the plight of my constituents
regarding this other issue is being ignored?'')

Second, an important characteristic of persuasive language is that 
speakers may spend more time presenting evidence in support of their positions
(or attacking the evidence presented by others) 
 than directly stating their attitudes.  An extreme example will
illustrate the problems involved. Consider a speech that 
describes the U.S. flag as 
deeply inspirational, 
and thus
contains only positive language.  
If the bill under discussion is
a proposed flag-burning ban, then the speech is {\em supportive}; but
if the bill under discussion is aimed at rescinding an existing flag-burning
ban, the speech may represent {\em opposition} to the legislation.  
Given the 
current state of the art in sentiment analysis, it is doubtful that one 
could determine the (probably topic-specific) relationship between 
presented evidence and speaker opinion.

\begin{table*}[th]
\begin{center}
\begin{tabular}{|l||r|rrl|} \hline
                      &total &train &test &development \\ \hline
\spunits              &3857  &2740     &860  &257 \\
\debates              &53    &38       &10   &5 \\
average number of \spunits per \debate  &72.8  &72.1  &86.0  &51.4  \\
average number of speakers per \debate  &32.1  &30.9    &41.1   &22.6   \\ \hline
\end{tabular}
\caption{\label{tab:data}
Corpus statistics. }
\end{center}
\end{table*}

\paragraph{Qualitative summary of results}

The above difficulties underscore the importance of enhancing standard
classification techniques with new information sources that 
promise to improve accuracy, 
such as inter-document relationships between the documents to be
labeled.
In this paper, we demonstrate that the incorporation of agreement modeling 
can provide substantial improvements over 
the application of
support vector machines
(SVMs) in isolation, which 
represents the state of the art in the individual classification 
of documents.  
The enhanced accuracies are obtained
via a fairly primitive automatically-acquired 
``agreement detector'' and a
conceptually simple method for integrating isolated-document and
agreement-based information.  We thus view our results as
demonstrating the potentially large benefits of exploiting 
sentiment-related discourse-segment relationships in
sentiment-analysis tasks.


\section{Corpus}
\label{sec:data}

This section outlines the main steps of the process by which we
created 
our corpus (download site:
www.cs.cornell.edu/home/llee/data/convote.html).

\medskip

GovTrack 
(\mbox{http://govtrack.us})
is an independent website 
run by Joshua Tauberer
that
collects publicly available
data on 
the legislative and fund-raising activities of U.S. congresspeople.  Due to its
extensive cross-referencing and collating of information, it 
was nominated for a 2006 ``Webby'' award.
A crucial characteristic of GovTrack from our point of
view is that the information is provided in a very convenient format;
for instance, the floor-debate transcripts are broken into
separate HTML files according to the subject of the debate, so we
can trivially derive long sequences of speeches guaranteed to cover the
same topic.

We extracted from GovTrack all available transcripts of U.S.  floor
debates in the House of Representatives for the year 2005 (3268 pages
of transcripts in total), together with voting records for all
roll-call
votes during that year.
We concentrated on \debates regarding ``controversial'' bills (ones in which the
losing side generated at least 20\% of the speeches) because
these debates should presumably exhibit more interesting
discourse structure.

Each 
debate consists of a series of {\em \spunits}, where each 
segment is a
sequence of uninterrupted utterances by a single speaker.
Since 
\spunits represent
natural  discourse units, we 
treat them as the basic 
unit to be classified.
Each \spunit was labeled by the vote (\y or \n) cast for the proposed bill
by the person who uttered the \spunit.

We automatically discarded those \spunits belonging to a class of 
formulaic, generally one-sentence utterances 
focused 
on the yielding of time on the house floor (for example, 
``Madam Speaker, I am pleased to yield 5 minutes to the gentleman from Massachusetts''), as such \spunits are
clearly off-topic.
We also removed \spunits 
containing the term ``amendment'', since we found 
during initial inspection that
 these speeches 
generally reflect a speaker's opinion on an amendment, and this
opinion may 
differ from the speaker's opinion 
on the underlying bill under discussion.

We randomly split the data into training, test, and development
(parameter-tuning) sets 
representing
roughly 70\%, 20\%, and 10\% of our data, respectively (see Table \ref{tab:data}).  
The \spunits remained grouped by \debate, with 38 
\debates
assigned to the training set, 
10 to the test set, and 5 to the development set; 
we 
require that the \spunits from an individual \debate
all appear in the same 
set because our goal is to
examine classification of \spunits in the context of the
surrounding discussion.


\section{Method}
\label{sec:method}

The support/oppose 
classification 
problem can be approached through 
the use of standard classifiers such as support vector machines (SVMs), 
which consider each text unit in isolation.
As discussed in Section \ref{sec:intro}, however, the conversational
nature of our data implies the existence of various relationships that
can be exploited 
to improve cumulative classification accuracy for 
\spunits belonging to the same \debate.
Our classification framework,
directly inspired by \newcite{Blum+Chawla:01a},
 integrates both perspectives, optimizing its
labeling of \spunits based on 
both individual 
\spunith classification scores and preferences for groups of \spunits to 
receive the same label.
In this section, we discuss 
the specific classification framework that we adopt
and the set of mechanisms that we propose for  
modeling specific types of relationships.

\subsection{Classification framework}
\label{sec:method:graph}

Let $\spvar_1,\spvar_2,\ldots, \spvar_{\numspunits}$ be the sequence
of \spunits within a given debate, and let $\yclass$ and $\nclass$ stand
for the ``yea'' and ``nay'' class, respectively. Assume we have a
non-negative function $\indscore(\spvar,\class)$ indicating the
degree of preference that an individual-document classifier, such as an SVM,
has for placing \spunith $\spvar$ in class $\class$.  Also, assume that 
some pairs of \spunits have {\em
weighted links} between them, where the non-negative {\em strength} (weight)
$\strength(\link)$ for a link $\link$ indicates the degree to which it
is preferable that the linked \spunits receive the same label.
Then, any class assignment $\classfn = \classfn(\spvar_1),
\classfn(\spvar_2),\ldots, \classfn(\spvar_{\numspunits})$
can be assigned a {\em cost}
$$
\sum_{\spvar}
\indscore(\spvar,\overline{\classfn}(\spvar)) +
\sum_{\spvar,\spvar':\: \classfn(\spvar) \neq \classfn(\spvar')}
\,
  \sum_{\ell\; {\rm between}\,  {\spvar, \spvar'} }\strength(\link),
$$
where $\overline{\classfn}(\spvar)$ is the ``opposite'' class from ${\classfn}(\spvar)$.
A {\em minimum-cost} assignment thus represents an optimum way to classify the
\spunits 
so that 
each one tends not to be put into the class that the
individual-document classifier disprefers, but at the same time, highly associated
\spunits tend not to be put in different classes.

As has been previously observed and exploited in the NLP literature
\cite{Pang+Lee:04a,Agarwal+Bhattacharyya:05a,Barzilay+Lapata:05a},
the above optimization function, unlike many others that have been proposed
for 
graph or set partitioning, can be solved {\em exactly} in
an 
provably
efficient manner via methods for finding minimum cuts in
graphs.  In our view, the contribution of our work is the examination
of new types of relationships, not the method by which such
relationships are incorporated into the classification decision.

\subsection{Classifying \spunits in isolation}
\newcommand{\svmlight}{${\rm  SVM}^{light}$\xspace}
In our experiments, we 
employed
the well-known 
classifier 
\svmlight
 to obtain individual-document 
classification scores, treating $\yclass$ as the positive 
class and using plain unigrams as features.\footnote{\svmlight is available at
svmlight.joachims.org.  
Default parameters were
used,
although experimentation with different parameter settings is an
important direction for future work
\cite{Daelemans+Hoste:02a,Munson+Cardie+Caruana:05a}.}
Following standard practice in sentiment analysis \cite{Pang+Lee+Vaithyanathan:02a}, the input to \svmlight consisted 
of normalized presence-of-feature (rather
  than frequency-of-feature) vectors.  The $\indscore$ value for each \spunit
 $\spvar$ 
was based on
the signed distance $\refconf$ from the vector representing
 $\spvar$ to the trained SVM decision 
plane:
$$
\indscore(\spvar,\yclass) \stackrel{{\rm def}}{=} 
\begin{cases}
1 & \refconf > 2\refstd; \cr
\left(1 + \frac{\refconf}{2\refstd}\right)/2 & 
\vert \refconf \vert \leq 2\refstd;
\cr
0 & \refconf < -2\refstd
\end{cases}
$$
where $\refstd$ is the standard deviation of  $\refconfvar(\spvar)$ over all
\spunits $\spvar$ in the
debate in question,
and $\indscore(\spvar,\nclass) \stackrel{{\rm def}}{=} 1 - \indscore(\spvar,\yclass)$.

We now turn to the more interesting problem of representing the
preferences that
\spunits may have for being assigned to the same class.

\subsection{Relationships between \spunits}
\label{sec:method:relationship}

A wide range of relationships between text segments 
can be modeled as positive-strength links.
Here we discuss two types of constraints that are considered in this work.

\paragraph{Same-speaker constraints:} 
In Congressional debates and in general 
social-discourse contexts, 
a single speaker may make 
a number of comments regarding 
a
topic.  
It is reasonable to expect that in many settings, the participants in
a discussion may be convinced
to change their opinions midway through a debate.
Hence,
in the general case we wish to be able 
to express
``soft'' preferences for all of an author's statements
to receive the same label, 
where
the strengths of such constraints could, 
for instance, vary according to the time elapsed between the statements.
Weighted links are an appropriate means to express such variation.

However, if
we assume that most speakers do not change their positions in the 
course of a discussion, we can conclude that all 
comments made by the same speaker
must
receive the same label.
This assumption
holds by fiat for the ground-truth labels in our dataset because these
labels were derived from the single vote cast by the speaker on the
bill being discussed.\footnote{
We are attempting to
determine whether a \spunit represents support or not.  This differs
from the problem of determining what the speaker's actual opinion is,
a problem that, as an anonymous reviewer 
put it, is complicated
by ``grandstanding, backroom deals, or, more innocently, plain change
of mind 
(`I voted for it before I voted against it')''.}
We can implement this assumption
via links whose weights are essentially infinite.  
Although
one can also implement this assumption via concatenation of
same-speaker \spunits (see
Section \ref{sec:eval:global}), we view the fact that our graph-based
framework  incorporates both hard and soft constraints in a principled
fashion as an
advantage of our approach.

\paragraph{Different-speaker agreements} In House discourse, it is common for one 
speaker to make reference to another
in the context of an
agreement or disagreement over the topic of discussion.
The systematic identification of instances of 
agreement 
can, as we have discussed, 
be a powerful tool 
for the development of intelligently
selected weights
for links between \spunits.

The problem of agreement identification 
can be decomposed into two
sub-problems: identifying references and their 
targets, and deciding whether
each reference represents an instance of agreement.
In our case, the first task is straightforward
because we focused solely on by-name references.\footnote {One subtlety
  is that for the purposes of mining agreement cues 
(but {\em not} for
  evaluating overall support/oppose classification accuracy), we 
temporarily
re-inserted into
our dataset previously filtered \spunits containing the term
``yield'', since the yielding of time on the House floor typically
indicates agreement even though the yield statements contain little
relevant text on their own.} 
Hence, we will now concentrate on the second, more interesting task.

We approach the problem of classifying references by 
representing each reference with a word-presence vector derived from a 
window of text surrounding 
the reference.\footnote{
We found good 
development-set performance using
the 30 tokens before,
20 tokens after,
and the name itself.}  
In the training set, we classify each reference connecting two speakers 
with a positive or negative label 
depending on whether the two
voted the same way on the bill under discussion\footnote{Since we are 
concerned with references that potentially represent relationships 
between \spunits, we ignore
references for which the target of the reference did not speak in the
\debate in which the reference was made.}.  These labels are then used to
train an SVM classifier, the output of which is subsequently used to
create weights
on {\em agreement links} in the test set as follows.

 Let $\agrconf$ denote the distance from the vector representing
 reference $\refvar$ to the agreement-detector SVM's decision plane,
 and let
$\agrstd$ be the standard deviation of  $\agrconfvar(\refvar)$ over all
 references in the \debate in question.
We then define the strength $\refscore$ of the {\em agreement link} corresponding
to the reference
as:
$$
\refscore(\refvar) \stackrel{{\rm def}}{=} \begin{cases}
0           & \agrconf < \refthres; \cr
\refcoef \cdot \agrconf/ 4\agrstd & \refthres \le \agrconf \le 4\agrstd; \cr
\refcoef & \agrconf > 4\agrstd.
\end{cases}
$$
The free parameter $\refcoef$ specifies the relative importance of the
$\refscore$ scores.
The threshold $\refthres$ controls 
the precision of
the agreement
links,
in that values of  $\refthres$ greater than zero mean that greater
confidence is required before an agreement link can be added.
\footnote{Our implementation
puts a link between
just one arbitrary pair of \spunits among all those uttered by a given
pair of
apparently agreeing speakers.  The ``infinite-weight'' same-speaker
links propagate the agreement information to all other such pairs.
}


\section{Evaluation}
\label{sec:eval}

This section presents experiments testing the utility  of 
using
\spunith relationships,
evaluating against a number of baselines.
All reported results use values for the
free parameter $\refcoef$ 
derived via tuning on the development set. 
In the tables, {\bf boldface} indicates the development- and
test-set results for the {\em development-set-optimal} parameter
settings, as one would make algorithmic choices based on
development-set performance.

\subsection{Preliminaries: Reference classification}
\label{sec:eval:agr}

Recall that to gather inter-speaker agreement information, the
strategy employed in this paper is to classify 
by-name 
references to other speakers as to whether they
indicate agreement or not.

\begin{table}[t]
\begin{tabular}{|l|cc|}
 \hline
\parbox[b]{4.5cm}{\mbox{Agreement classifier} \\ \hspace*{.2cm}(``{reference}$\Rightarrow$agreement?'')}      &
\parbox[b]{1cm}{Devel. set}  &\parbox[b]{1cm}{Test set} \\
\hline\hline

\resmajoritybl & 81.51 & 80.26 \\
\agrSVMtrainnoamend & 84.25 & 81.07 \\
\agrSVMtrainwamend & \bestdev{86.99} & \testbestdev{80.10} \\
\hline

\end{tabular}
\caption{\label{tab:agr} 
Agreement-classifier accuracy, in percent.  ``Amdmts''=``\spunits containing the
word {`amendment'}''.
Recall that boldface indicates results for
  development-set-optimal settings.
}
\end{table}

To train 
our agreement classifier,
we
experimented with undoing the deletion of amendment-related \spunits
in the training set.  Note that such \spunits were {\em never}
included in the development or test set, since, as discussed in 
Section
 \ref{sec:data}, their labels are probably noisy; however, including them in
the {\em training} set allows the classifier to examine more instances even
though some of them are labeled incorrectly.  As Table
\ref{tab:agr} shows, using more, if noisy, data yields better
agreement-classification results on the {development} set, and so we use
that policy in all subsequent experiments.
\footnote{
Unfortunately, this policy leads to inferior {\em test-set}
agreement classification.  Section \ref{sec:eval:devtest}
contains further discussion.
}

An important observation is that precision may be more important than
accuracy in deciding which agreement links to add: false positives
with respect to 
agreement
can cause \spunits to be
incorrectly assigned the same label, whereas false negatives mean only
that agreement-based information about other \spunits is not employed.  As described above, we
can raise agreement precision by 
increasing the threshold $\refthres$,
which specifies 
the required confidence for the addition of an agreement link.
Indeed, Table \ref{tab:agr-highprec} shows that we can 
improve 
agreement precision
by 
setting  $\refthres$  to the (positive) mean agreement score $\refmean$ assigned by the SVM
agreement-classifier over all references in the given
{\debate}\footnote{We elected not to explicitly tune the value of
$\refthres$ in order to minimize the number of free parameters to deal
with.}.  However, this comes at the cost of
greatly reducing agreement accuracy (development: 64.38\%; test:
\fix{66.18}{66.34}\%) due to 
lowered recall levels.  Whether or not better
\spunith classification 
is ultimately achieved 
is discussed
in the next sections.

\begin{table}[t]
\begin{tabular}{|l|cc|}
 \hline
{{Agreement classifier}}      &
\multicolumn{2}{|c|}{Precision (in percent):} \\
& 
{Devel. set}  &
{Test set} \\
\hline\hline

\useposref & 86.23 & 82.55 \\
\useavgref & \bestdev{89.41} & \testbestdev{\fix{88.47}{88.50}} \\
\hline

\end{tabular}
\caption{\label{tab:agr-highprec} 
Agreement-classifier precision.
}
\end{table}

\subsection{\Localsvmsetting}
\label{sec:eval:local}

\paragraph{Baselines} The first two data rows of Table
\ref{tab:results-local} depict baseline performance results.  
The 
\reswordbl
 baseline is meant to explore whether the
 \spunith classification task can be reduced to simple
lexical checks.  Specifically, this method uses the signed difference
between the number of words containing the stem ``support'' and the
number of words containing the stem ``oppos'' (returning the
majority class if the difference is 0).  No better than 62.67\%
test-set accuracy is obtained by either baseline.

\providecommand{\per}{}
\begin{table}[t]
\begin{center}
\begin{tabular}{|l|cc|} \hline
\parbox[b]{4.3cm}{Support/oppose classifer \\ \hspace*{.2cm}(``{\spunit}$\Rightarrow$yea?'')}      & \parbox[b]{1cm}{Devel. set}  &\parbox[b]{1cm}{Test set} \\ \hline\hline
\resmajoritybl  &54.09\per        &58.37\per \\
\reswordbl      &59.14\per        &62.67\per \\ \hline
\ressvmbl [{\spunit}]      &70.04\per        &66.05\per \\
\resspeakerlink &79.77\per        &67.21\per \\ 
\resspeakerlink $\ldots$  & & \\
{~~~} \resfullref        &\bestdev{89.11\per}        &\testbestdev{70.81\per} \\ 
{~~~} \reslimitedref     &87.94\per        & \fix{71.16}{70.81\per} \\ \hline
\end{tabular}
\caption{\label{tab:results-local}
\Localsvmsetting accuracy, in percent.}
\end{center}
\end{table}

\providecommand{\per}{}
\begin{table}[t]
\begin{center}
\begin{tabular}{|l|cc|} \hline
\parbox[b]{4.3cm}{Support/oppose classifer \\ \hspace*{.2cm}(``{\spunit}$\Rightarrow$yea?'')}      & \parbox[b]{1cm}{Devel. set}  &\parbox[b]{1cm}{Test set} \\ \hline\hline
\ressvmbl [speaker]      &71.60\per        &70.00\per \\
SVM + \refs $\ldots$ & & \\
{~~~} with \useposref
     &\bestdev{88.72}\per        &\testbestdev{71.28}\per \\ 
{~~~} with \useavgref
 &84.44\per        &\fix{76.05}{76.16\per} \\ 
\hline
\end{tabular}
\caption{\label{tab:results-global}
\Globalsvmsetting accuracy,
in percent.  
Here,
the
initial SVM is run on the concatenation of all of a given speaker's
\spunits,
but 
the results 
are computed over \spunits
(not speakers),
so that 
they can be
compared to those in Table \ref{tab:results-local}.
}
\end{center}
\end{table}

\paragraph{Using relationship information}
Applying an SVM to classify each \spunit in isolation leads to clear
improvements over the two baseline methods, as demonstrated in 
Table \ref{tab:results-local}.
When we impose the constraint that all \spunits uttered by the same speaker
receive the same label
via ``same-speaker links'', 
both test-set
and development-set accuracy increase even more, in the latter case
quite substantially so.

The last two lines of Table \ref{tab:results-local} show that the best
results are obtained by incorporating agreement information as well.
The highest test-set result, \fix{71.16}{70.81}\%, is obtained by using \fix{a high-precision
threshold to determine which agreement links to add}{agreement links,
  with the   \useposref\xspace policy performing better than 
 \useavgref\xspace on the
  development set}.  \fix{While the
development-set results would induce us to utilize the standard threshold
value of 0, which is sub-optimal on the test set, the
\mbox{\useposref}
agreement-link
policy still achieves noticeable improvement 
over  not using  agreement links
(test set: 70.81\% vs. 67.21\%).}

\subsection{\Globalsvmsetting}
\label{sec:eval:global}

We 
use \spunits as the  unit of classification 
because
they represent natural discourse 
units.
As a  consequence, 
we 
are able to
exploit 
relationships 
at the \spunith
level.  
However, it is interesting to consider whether we really need to
consider relationships specifically between \spunits themselves, or whether it
suffices to simply consider relationships between the {\em speakers} of
the \spunits.
In particular, as an alternative to using same-speaker links, 
we tried a {\em \globalclass} approach wherein 
the way we determine
the initial
individual-document classification score for 
each 
\spunit uttered by a
person $p$ in a given debate 
is to run
an SVM 
on the
concatenation of
{\em all} of $p$'s \spunits
within that debate.
(We also ensure that agreement-link information is propagated from
\spunith 
to speaker pairs.)

How 
does the use of 
same-speaker links compare to the concatenation of
each speaker's \spunits?   
Tables  \ref{tab:results-local} and \ref{tab:results-global} 
show that, not surprisingly, 
the SVM individual-document
classifier works better on the concatenated \spunits than on the
\spunits in isolation.
However, the effect on overall
classification accuracy is less clear: the development set favors
same-speaker links over concatenation, 
while
the test set does not.

But we stress that
the most important observation we can make from 
 Table \ref{tab:results-global} is  that once again, the
addition of agreement information leads to substantial improvements in
accuracy.

\subsection{``Hard'' agreement constraints}  Recall that in 
in our experiments, 
we created 
finite-weight
agreement links, so that \spunits appearing in pairs flagged by our (imperfect)
agreement detector can potentially receive different labels.  We also
experimented with 
{\em forcing} such \spunits to receive the same label,
either through infinite-weight agreement links or through a \spunith
concatenation strategy similar to that described in the previous
subsection.  Both strategies resulted in clear degradation in
performance on both the development and test sets, a finding that 
validates our encoding of agreement information as ``soft'' preferences.

\subsection{On the development/test set split}
\label{sec:eval:devtest}

We have seen several cases in which the
method that performs best on the development set does not yield the
best test-set performance.  However, we felt that it would be
illegitimate to change the train/development/test 
sets 
in a post hoc
fashion, that is, after seeing the experimental results.  

Moreover, and crucially, it is very clear that using agreement
information, encoded as preferences within our graph-based approach
rather than as hard constraints, yields substantial improvements on
both the development and test set; this, we believe, is our most
important finding.


\section{Related work}

\paragraph{Politically-oriented text}  
Sentiment analysis has specifically been proposed as a key enabling
technology in eRulemaking, allowing the  automatic analysis of
the opinions that people submit
\cite{Shulman+al:05a,Cardie+al:06a,Kwon+Shulman+Hovy:06a}.
There has also been work focused upon
determining the political leaning (e.g.,
``liberal'' vs. ``conservative'') of a document or 
author, 
where most
previously-proposed methods make no direct use of relationships
between the documents to be classified (the ``unlabeled'' texts)
\cite{Laver+Benoit+Garry:03a,Efron:04a,Mullen+Malouf:06a}.
An exception is \newcite{Grefenstette+Qu+Shanahan+Evans:04a}, who experimented with
  determining the political orientation of websites essentially by
  classifying the concatenation of all the documents found on that site.

Others have applied 
the NLP technologies of near-duplicate detection and
topic-based text categorization to politically oriented text
\cite{Yang+Callan:05a,Purpura+Hillard:06a}.

\paragraph{Detecting agreement} We used a simple method to learn to
identify 
cross-speaker
references 
indicating
agreement.  More sophisticated approaches
have been proposed
\cite{Hillard+Ostendorf+Shriberg:03a}, including an extension that, in
an interesting reversal of our problem, makes use of
sentiment-polarity indicators within 
\spunits \cite{Galley+McKeown+Hirschberg+Shriberg:04a}.  
Also relevant is work on the general problems of dialog-act tagging
\cite{Stolcke+al:00a},  citation analysis
\cite{Lehnert+Cardie+Riloff:90}, and computational rhetorical analysis \cite{Marcu:00a,Teufel+Moens:02a}.

We currently do not have an efficient means to encode {\em
disagreement} information as hard constraints; we plan to
investigate incorporating
such information in future work.

\paragraph{Relationships between the unlabeled items}  
\newcite{Carvalho+Cohen:05a} consider sequential relations between different
types of emails (e.g.,
between requests and satisfactions thereof) to classify
messages, and thus also explicitly exploit the structure of 
conversations.

Previous
sentiment-analysis work in different domains
has considered inter-document similarity
\cite{Agarwal+Bhattacharyya:05a,Pang+Lee:05a,Goldberg+Zhu:06a} or 
explicit inter-document references in the form of hyperlinks  \cite{Agrawal+al:03a}.

Notable early papers on graph-based semi-supervised learning include
\newcite{Blum+Chawla:01a}, \newcite{Bansal+Blum+Chawla:02a}, \newcite{Kondor+Lafferty:02a}, and
\newcite{Joachims:03a}.  \newcite{Zhu:semi-survey} maintains a survey
of this area.

Recently, several alternative, often quite sophisticated approaches to {\em collective
  classification} have been proposed
\cite{Neville+Jensen:00a,Lafferty+McCallum+Pereira:01a,Getoor+al:02a,Taskar+al:02a,Taskar+Guestrin+Koller:03a,Taskar+Chatalbashev+Koller:04a,McCallum+Wellner:04a}.
It would be interesting to investigate the application of such methods
to our problem.  However,  we also believe that our approach has
important advantages, including  conceptual simplicity and 
the fact that it
is based on an  underlying optimization problem that is provably
and 
in practice
easy to solve.


\section{Conclusion and future work}
\label{sec:conc}

In this study, we focused on very general 
types
of cross-document
classification preferences, 
utilizing constraints based only on speaker
identity and on direct textual references between statements.  
We showed that the integration of 
even very limited information regarding
inter-document relationships
can significantly increase the accuracy
of support/opposition classification.

The simple constraints 
modeled in our study, however, represent 
just a small 
portion 
of the rich network of relationships that connect
statements 
and speakers
across the political universe and in the wider realm of
opinionated social discourse.  
One intriguing possibility is to take advantage of (readily identifiable)
information regarding interpersonal relationships, making use of 
speaker/author affiliations, positions
within a social hierarchy, and so on.
Or, we could even  attempt to model relationships
between
topics or concepts, 
in a kind of extension of collaborative filtering.
For example, 
perhaps we could infer that 
two
speakers sharing a common opinion on evolutionary biologist Richard
Dawkins (a.k.a. ``Darwin's rottweiler'') will be likely to agree
in a debate centered on Intelligent Design.  While such functionality is
well beyond the scope of our current study, we are optimistic that we
can develop methods to exploit
additional types of relationships in future work.


\paragraph*{Acknowledgments}  We thank 
Claire Cardie, 
Jon Kleinberg, Michael Macy, Andrew
Myers, 
and the six anonymous EMNLP referees for valuable discussions
and comments.  We also thank Reviewer 1 for generously providing
additional {\em post hoc} feedback, and the EMNLP chairs Eric Gaussier
and Dan Jurafsky for facilitating the process (as well as for
allowing authors an extra proceedings page$\ldots$).
This paper is based upon work supported in part by the National
Science Foundation under grant no. IIS-0329064 
{\color{red}{[Addition:], an Alfred P. Sloan
Research Fellowship, and Google Anita Borg Memorial Scholarship funds}}. Any opinions,
findings, and conclusions or recommendations expressed are those of
the authors and do not necessarily reflect the views or official
policies, either expressed or implied, of any sponsoring institutions,
the U.S. government, or any other entity.

\vspace*{-.1in}

\bibliographystyle{fullname}

\end{document}